\documentclass[runningheads]{llncs}

\usepackage{eccv}

\usepackage{eccvabbrv}

\usepackage{graphicx}
\usepackage{booktabs}
\usepackage[numbers,sort&compress]{natbib}

\usepackage{multirow}
\usepackage{tabularx}

\usepackage[accsupp]{axessibility}  

\usepackage{hyperref}

\usepackage{orcidlink}

\begin{document}

\title{TreeGaussian: Tree-Guided Cascaded Contrastive Learning for Hierarchical Consistent 3D Gaussian Scene Segmentation and Understanding} 

\titlerunning{TreeGaussian}

\newcommand{\xz}[1]{{\color{blue}[xz: #1]}}

\author{Jingbin You\inst{1,2} \and
 Zehao Li\inst{1,2} \and
 Hao Jiang\inst{1,2} \and
 Xinzhu Ma\inst{3} \and
 Shuqin Gao\inst{1,2} \and
 Honglong Zhao\inst{1,2} \and
 Congcong Zheng\inst{1,2} \and
 Tianlu Mao\inst{1,2} \and
 Feng Dai\inst{1,2} \and
 Yucheng Zhang\inst{1,2} \and
 Zhaoqi Wang\inst{1,2} 
 }

\authorrunning{You et al.}

\institute{
Institute of Computing Technology, Chinese Academy of Sciences, ICT \and
University of Chinese Academy of Sciences, UCAS \and
Beihang University, BUAA \\
 }

\maketitle

\begin{abstract}
  3D Gaussian Splatting (3DGS) has emerged as a real-time, differentiable representation for neural scene understanding. However, existing 3DGS-based methods struggle to represent hierarchical 3D semantic structures and capture whole–part relationships in complex scenes. Moreover, dense pairwise comparisons and inconsistent hierarchical labels from 2D priors hinder feature learning, resulting in suboptimal segmentation. To address these limitations, we introduce TreeGaussian, a tree-guided cascaded contrastive learning framework that explicitly models hierarchical semantic relationships and reduces redundancy in contrastive supervision. By constructing a multi-level object tree, TreeGaussian enables structured learning across object-part hierarchies. In addition, we propose a two-stage cascaded contrastive learning strategy that progressively refines feature representations from global to local, mitigating saturation and stabilizing training. A Consistent Segmentation Detection (CSD) mechanism and a graph-based denoising module are further introduced to align segmentation modes across views while suppressing unstable Gaussian points, enhancing segmentation consistency and quality. Extensive experiments, including open-vocabulary 3D object selection, 3D point cloud understanding, and ablation studies, demonstrate the effectiveness and robustness of our approach.
\end{abstract}

\section{Introduction}
\label{sec:intro}

\begin{figure}[t]
\centering\includegraphics[width=0.9\linewidth]{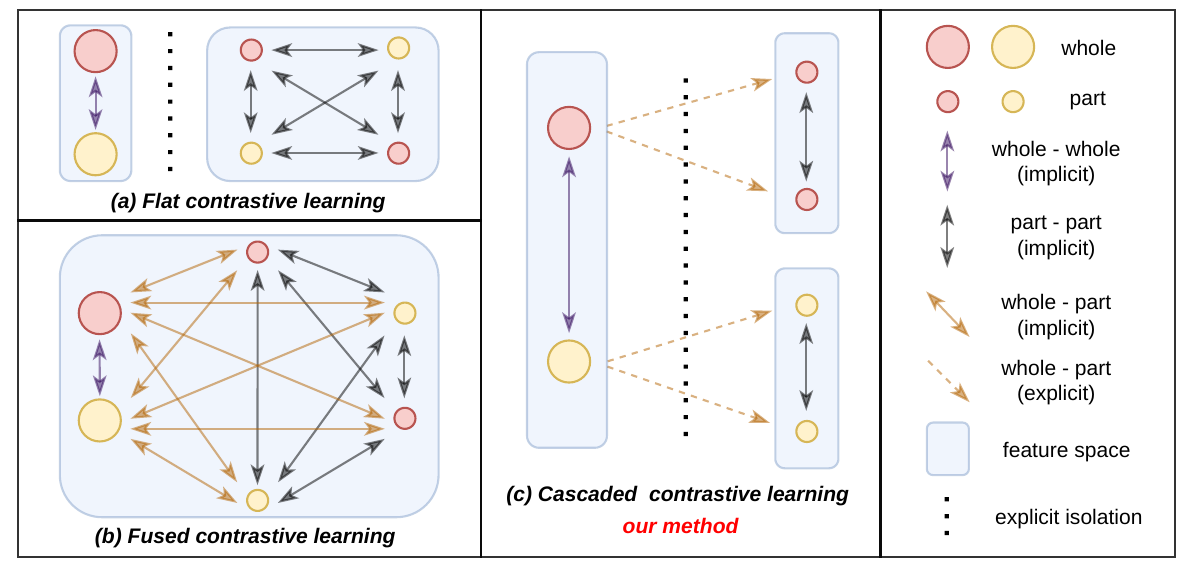}
\caption{Motivation illustration.
\textit{(a) Flat contrastive learning} isolates feature spaces for object wholes and parts, limiting their hierarchical interaction. 
\textit{(b) Fused contrastive learning} merges feature spaces but suffers from oversaturation and instability due to dense pairwise comparisons. 
\textit{(c) Cascaded contrastive learning (our method)} preserves semantic hierarchy while minimizing contrastive redundancy, enabling efficient and stable optimization. 
Solid arrows indicate implicit contrastive interactions among clusters; dashed arrows denote explicit hierarchical indexing relationships.
}
\label{fig:comparsion_constrasted_learning}
\end{figure}

3D scene understanding is foundational to high-level perception, with broad applications in virtual reality \citep{dai2017scannet, mccormac2017semanticfusion}, autonomous navigation \citep{mur2017orb,humblot2022navigation}, robotic interaction \citep{saxena2008robotic,zheng2024gaussiangrasper,li2024object}, \etc 
A central determinant of this capability is the choice of 3D representation, which decides how scene geometry and appearance are stored and learned.
Recently, 3D Gaussian Splatting (3DGS) \citep{kerbl20233d} has emerged a strong and efficient representation. 
It models scenes with compact Gaussian primitives, enabling high-quality reconstruction at low computational cost.
These strengths make 3DGS a promising foundation  for semantic tasks such as open-vocabulary segmentation and scene understanding.

Despite the fact that 3DGS excels in geometry and appearance modeling, it lacks native semantic representation.
To bridge this gap, recent studies have explored the integration of image-derived features into Gaussian representations.
Early approaches \citep{qin2024langsplat,LangEmbedGaussian} embed high-dimensional language features from vision–language models \citep{radford2021learning,caron2021emerging} into 3D scenes via compression to support open-vocabulary segmentation.
However, such compression inevitably sacrifices semantic richness.
To mitigate this issue, recent methods \citep{wuopengaussian, li2024instancegaussian, ying2024omniseg3d} have adopt contrastive learning \citep{wang2020understanding} with 2D segmentation priors \citep{segmentAnything} to supervise the low-dimensional instance features within Gaussians. 
While contrastive learning offers a more flexible form of supervision, it introduces a new challenge in organizing semantic information effectively. 
Specifically, these methods struggle to accurately represent hierarchical 3D semantic structures and effectively capture the structured relationships between object wholes and parts.

This challenge is illustrated in Figure~\ref{fig:comparsion_constrasted_learning}, which compares representative contrastive learning methods and highlights their difficulty in building structured and consistent semantic representations. Flat contrastive learning methods such as OpenGaussian \citep{wuopengaussian} and InstanceGaussian \citep{li2024instancegaussian} treat semantic entities as isolated instance features. This separation between whole objects and parts leads to a lack of hierarchical structure, which prevents the model from learning structured semantic relationships. Fusion-based methods such as OmniSeg3D-GS \citep{ying2024omniseg3d} attempt to combine feature spaces across levels. However, in complex scenes with many small objects, the number of feature pairs becomes overwhelming, leading to low-dimensional feature space saturation and unstable optimization. Moreover, inconsistent segmentation scales across views introduce conflicting supervision signals, preventing hierarchically consistent multi-view segmentation in both flat and fused methods. These limitations highlight the need for a more structured and scalable contrastive framework capable of capturing semantic hierarchies and ensuring consistent supervision across complex scenes.

Motivated by these constraints, we propose TreeGaussian, a tree-guided cascaded contrastive learning framework for Gaussian-based open-vocabulary segmentation. TreeGaussian introduces a hierarchical object tree to organize Gaussians across multiple levels of abstraction, explicitly modeling structured relationships between object parts and wholes. This structure reduces semantic redundancy and enables more meaningful contrastive supervision. Furthermore, we design a two-stage cascaded contrastive learning strategy: a global learning stage that enhances discrimination across coarse-grained semantic clusters and a local learning stage that refines features within fine-grained clusters. This progressive learning process improves both the quality of the representation and the stability of the training. Additionally, to prevent multi-view inconsistent supervision, we introduce a Consistent Segmentation Detection mechanism. It dynamically selects the appropriate contrastive component for each view and aligns segmentation modes to resolve over-segmentation and under-segmentation, ensuring consistent segmentation scales across multiple views. To further mitigate the impact of noisy supervision, we incorporate a graph-based denoising module that identifies and removes semantically unstable Gaussian points by evaluating their spatial and feature-level consistency. Through comprehensive experiments, including open-vocabulary 3D object selection, 3D point cloud understanding, and ablation studies, we demonstrate the robustness and reliability of our proposed method. Our main contributions are as follows: 
\begin{itemize}
    \item We analyze key limitations of current contrastive learning frameworks for 3D Gaussian scene segmentation, including the lack of hierarchical semantic modeling, oversaturation of contrastive pairs, and inconsistent segmentation scales from multiple views. 
    \item We propose a tree-guided cascaded contrastive learning framework that leverages hierarchical semantic structures and progressive feature refinement to improve training stability and semantic representation quality. 
    \item We introduce a Consistent Segmentation Detection (CSD) mechanism to dynamically align segmentation modes and ensure segmentation consistency across views.
    \item We design a graph-based denoising module to suppress unstable Gaussian points, improving the quality of the final segmentation.
\end{itemize}

\section{Related Work}
\label{sec:Related Work}

\subsection{Neural Rendering}
3D scene representation plays a crucial role in scene understanding and directly influences the fidelity and efficiency of 3D reconstruction. Traditional 3D representations, such as voxels \citep{han2020occuseg, liu2019point, fridovich2022plenoxels}, meshes \citep{valentin2013mesh, schult2020dualconvmesh}, and point clouds \citep{yang2019learning, yi2019gspn, peng2023openscene, aliev2020neural}, often rely on non-differentiable rendering pipelines and fixed-category classifiers. These constraints limit their capacity for end-to-end optimization and generalization to novel concepts. 
Neural Radiance Fields (NeRF) \citep{mildenhall2021nerf} mark a milestone in neural rendering by representing scenes as continuous volumetric functions that enable photo-realistic view synthesis. However, NeRF-based methods \citep{kerr2023lerf, cen2023segment, kim2024garfield} require extensive sampling and computation, resulting in slow rendering and poor scalability to complex scenes.
In recent years, methods \citep{qin2024langsplat, zhou2024feature, LangEmbedGaussian, wuopengaussian} based on 3D Gaussian Splatting (3DGS) \citep{kerbl20233d} have demonstrated significant advantages in rendering and 3D representation. 3DGS is a technique for explicit data representation that efficiently and discretely encodes 3D data. The differentiable rasterization renderer \citep{lassner2021neuralRendering} used by 3DGS can greatly accelerate the rendering of novel views. Our approach is based on the explicit representation of 3DGS and focuses on detailed hierarchical segmentation and understanding of 3D scenes.

\subsection{3D Scene Segmentation and Understanding}

\noindent
\textbf{Language Distillation.} LERF \citep{kerr2023lerf} is the first method to distill CLIP \citep{radford2021learning}'s semantic information into the NeRF radiance field, enabling the execution of open-vocabulary language query tasks \citep{guadarrama2014open} within 3D scenes. LangSplat \citep{qin2024langsplat} integrates language with 3D Gaussian representations, aligning CLIP’s language features to pixel-level precision using SAM, and learns the hierarchical semantics defined by SAM \citep{segmentAnything} to address point ambiguity issues in 3D language field modeling. Feature3DGS \citep{zhou2024feature} inserts rendered feature maps into the feature space of 2D backbone models (such as SAM \citep{segmentAnything}, CLIP \citep{radford2021learning}, LSeg \citep{li2022lseg}, DINO \citep{caron2021dino}), utilizing the decoder of the backbone models to obtain segmentation results. N2F2 \citep{n2f2} employs nested feature representation \citep{kusupati2022nestedfeature}, using neural networks to learn the language features of various objects at different scales. However, directly distilling high-dimensional semantic features relies on an encoder or decoder, and during the dimensionality compression process, some semantic information may be lost. In contrast, our method utilizes contrastive learning strategy to optimize low-dimensional instance-level features, effectively avoiding the semantic loss caused by direct language distillation.



\noindent
\textbf{Contrastive Learning.}
Recent open-vocabulary 3D segmentation methods primarily rely on contrastive learning to optimize instance features within 3D representations. 
OpenGaussian~\citep{wuopengaussian} performs 2D-3D instance-level alignment to assign language embeddings to 3D Gaussian points.
InstanceGaussian~\citep{li2024instancegaussian} jointly learns appearance and instance features through a segment-first, aggregate-later paradigm. 
SAGA~\citep{cen2025segment} introduces a soft scale-gating mechanism to alleviate multi-granularity ambiguity.
Click-Gaussian~\citep{choi2024click} leverages secondary mask supervision to aggregate 2D masks into explicit 3D Gaussian representations.
GarField~\citep{kim2024garfield} utilizes scale-conditioned feature fields to aggregate masks at different scales into NeRF. 
Despite these advances, most existing approaches adopt flat semantic modeling and perform dense pairwise contrastive learning across instances, which can lead to feature space saturation and limited hierarchical discrimination in complex scenes.
OmniSeg3D~\citep{ying2024omniseg3d} uses fusion-based hierarchical contrastive learning to optimize consistent 3D hierarchical feature fields on the same view. Although it partially improves scale consistency, it does not explicitly decouple whole-object and part-level feature spaces, which may cause entangled representations across semantic levels.
In contrast, our method introduces a tree-guided cascaded contrastive framework that separates global whole-object learning from local part-level refinement, thereby reducing contrastive redundancy and enhancing hierarchical representation capacity.

\section{Method}
\label{sec:method}



In this section, we present our tree-guided cascaded learning framework for hierarchical 3D scene segmentation. 
We begin by constructing multi-view 2D object trees from SAM-generated masks to model hierarchical semantic relationships across scales (Sec.~\ref{sec:object_tree}). 
Then, we design a two-stage cascaded contrastive learning strategy that progressively optimizes Gaussian instance features from global to local levels (Sec.~\ref{sec:two_stage_contrastive}). 
To ensure stable optimization under multi-view variations, a Consistent Segmentation Detection (CSD) mechanism is introduced to dynamically balance contrastive supervision across views (Sec.~\ref{sec:csd}). 
Finally, a graph-based Gaussian denoising module further refines instance clusters by removing noisy points while preserving structural integrity (Sec.~\ref{sec:denoising}).

\subsection{Multi-View 2D Object-Tree Construction}
\label{sec:object_tree}

\begin{figure*}[t]
\centering\includegraphics[width=1.0\linewidth]{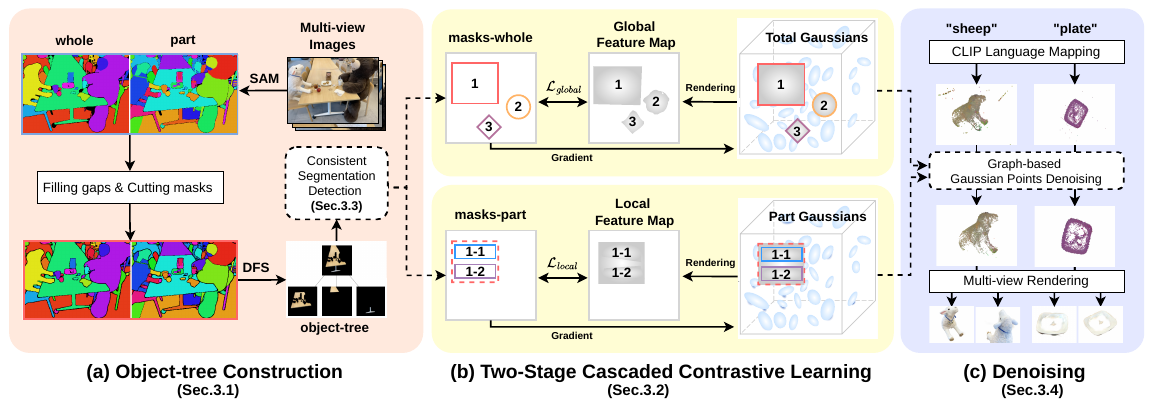}
\caption{Overview of our method. (a) Constructing object-tree from multi-view images using SAM to capture structured relationships between object parts and wholes. (b) Two-stage cascaded contrastive learning strategy to progressively optimize the instance feature of each Gaussian point. (c) Graph-based denoising is applied to each language-mapped instance cluster for improving the multi-view rendering quality.} 
\label{fig:pipeline}
\end{figure*}

As shown in Figure~\ref{fig:pipeline}(a), we leverage multi-level masks generated by SAM~\citep{segmentAnything} as our initial mask set to model hierarchical semantic relationships. 
However, the initial masks at different scales exhibit substantial overlaps and gaps, which hinder the construction of accurate hierarchical relationships. To enable clean hierarchical organization, we refine these masks through the following procedures: 

\begin{itemize}
  \item Filling Missing Masks: Gaps in the large-scale mask map are filled using complementary smaller-scale masks, while gaps in the small-scale mask map are filled using corresponding larger-scale masks.
  \item Edge Detection: An 8-neighborhood edge detector ([[1,1,1],[1,0,1],[1,1,1]]) is applied to set the edges of the multi-scale mask map to zero.
  \item Mask Cutting: Locations where the large-scale mask map is zero are also set to zero in the small-scale mask map to eliminate cross-scale conflicts and ensure strict spatial alignment for hierarchical consistency.
  \item Mask Filtering: Noisy mask fragments smaller than a predefined threshold are discarded to remove insignificant regions.
\end{itemize}

After these preprocessing steps, we perform a Depth First Search (DFS) starting from the mask map at the largest scale. This process generates multiple mask trees, which are then used to construct multi-view 2D instance trees for each object. Each child node mask is guaranteed to be fully contained within its corresponding parent node mask. 
More details of the multi-level mask preprocessing procedure are provided in the \textit{Supplementary Material (Figure S1)}.
The processed multi-view 2D instance trees are further employed to supervise 3D instance features, thereby enabling hierarchical semantic optimization.

\subsection{Two-Stage Cascaded Contrastive Learning}
\label{sec:two_stage_contrastive}

\noindent
\textbf{Feature Rendering.} 3DGS represents a scene with explicit 3D Gaussians and renders it via a differentiable rasterizer \citep{lassner2021neuralRendering}. Given multi-view images $\mathcal{I} = \{I_v\}_{v=1}^V$, it learns a set of Gaussians $\mathcal{G} = \{g_i\}_{i=1}^N$, where $V$ and $N$ represent the number of views and Gaussians. Each Gaussian $g_i$ is parameterized by a center position ${p}_i \in \mathbb{R}^3$, a scaling factor ${s}_i \in \mathbb{R}^3$, and a rotation encoded as a quaternion ${q}_i \in \mathbb{R}^4$. Additionally, each Gaussian has an opacity $o_i \in \mathbb{R}$ and a color $c_i$ represented by spherical harmonics coefficients \citep{fridovich2022sphericalHarmonics}. For rendering, Gaussians are projected onto the image plane according to the camera pose, and the pixel color $C$ is computed via alpha blending \citep{foley1996alphaBlending}:
\begin{equation}
\label{eq:1}
C = \sum_{i \in \mathcal{N}} c_i \alpha_i T_i,
\end{equation}
where $\alpha_i$ is the visibility weight derived from the projected 2D covariance, the opacity $o_i$, and the distance to the pixel center. The transmittance $T_i$ is computed recursively as $T_i = \prod_{j=1}^{i-1} (1 - \alpha_j)$, ensuring proper compositing order based on depth. To enable instance segmentation over the reconstructed scene, each 3D Gaussian $g_i$ is further augmented with a learnable low-dimensional instance feature $\boldsymbol{f}_i \in \mathbb{R}^D$, where $D$ denotes the feature dimensionality. This feature captures structural information associated with the corresponding region in the scene. The complete parameter set for each Gaussian becomes $g_i = \{{p}_i, {s}_i, {q}_i, o_i, c_i, \boldsymbol{f}_i\}$. Using the differentiable rasterizer, we render the instance feature map $\boldsymbol{F}$ for each pixel by aggregating the features $\boldsymbol{f}_i$ of the 3D Gaussians projected onto the pixel. This feature rendering process follows the same alpha blending strategy:
\begin{equation}
\boldsymbol{F} = \sum_{i \in \mathcal{N}} \boldsymbol{f}_i \alpha_i T_i.
\end{equation}
\noindent
\textbf{Global Contrastive Learning.} In the global learning stage, we employ decoupled global contrastive objectives to learn the discriminative features at the coarsest scale. Given a rendered global feature map $\boldsymbol{F}\in\mathbb{R}^{D\times H\times W}$ from total Gaussian points and a set of $m_1$ large-scale instance masks $\{\boldsymbol{M}_i^1\}_{i=1}^{m_1}$ obtained from the object tree, each region prototype is computed as $\overline{\boldsymbol{F}}_{i}=\left(\boldsymbol{M}_{i} \cdot \boldsymbol{F}\right) / \sum \boldsymbol{M}_{i} \in \mathbb{R}^{D} $. We optimize the global contrastive loss 
$
\mathcal{L}_{global} \;=\;\mathcal{L}_{pull}^1 + \mathcal{L}_{push}^1
$
with:
\begin{equation}
\mathcal{L}_{pull}^1
= \sum_{i=1}^{m_1} \sum_{h=1}^{H} \sum_{w=1}^{W} \, \boldsymbol{M}_{i,h,w}^1 \,\big\|\boldsymbol{F}_{:,h,w} - \overline{\boldsymbol{F}}_{i}\big\|^2,
\end{equation}
and:
\begin{equation}
    \mathcal{L}_{push}^1=\frac{1}{m_1(m_1-1)} \sum_{i=1}^{m_1} \sum_{j=1, j \neq i}^{m_1} \frac{1}{\left\|\overline{\boldsymbol{F}}_{i}-\overline{\boldsymbol{F}}_{j}\right\|^{2}},
\end{equation}
where $\mathcal{L}_{pull}^1$ aims to minimize the distance of the instance features within the same object and $\mathcal{L}_{push}^1$ aims to maximize the distance of the instance features from different objects. After convergence, K-means clustering \citep{macqueen1967kmeans} is applied, jointly considering the instance features $\boldsymbol{f}_i$ and spatial relationships ${p}_i$, to partition the scene into coarse object nodes for the next stage.


\noindent
\textbf{Local Contrastive Learning.} We introduce a cascaded local contrastive stage to overcome feature space saturation when many small object parts compete. For each coarse node, its associated part Gaussians are rendered as an independent subscene from the visible viewpoint. The rendered local feature map is converted to a binary region $\boldsymbol{B}\in\{0,1\}^{H\times W}$. Within this localized region, we generate $m_2$ finer-scale masks $\{\boldsymbol{M}_i^2\}_{i=1}^{m_2}$ from the object tree. Each region prototype is computed as $\overline{\boldsymbol{F}}_{i}=\left(
\boldsymbol{B}\cdot\boldsymbol{M}_{i} \cdot \boldsymbol{F}\right) / \sum \boldsymbol{B} \cdot \boldsymbol{M}_{i} \in \mathbb{R}^{D}$. We optimize the local contrastive loss 
$
\mathcal{L}_{local}
$
in combination with:
\begin{equation}
\mathcal{L}_{pull}^2
= \sum_{i=1}^{m_2} \sum_{h=1}^{H} \sum_{w=1}^{W}
   \boldsymbol{B}_{h,w}\,\boldsymbol{M}_{i,h,w}^2\;\big\|\boldsymbol{F}_{:,h,w} - \overline{\boldsymbol{F}}_{i}\big\|^2,
\end{equation}
and:
\begin{equation}
    \mathcal{L}_{push}^2=\frac{1}{m_2(m_2-1)} \sum_{i=1}^{m_2} \sum_{j=1, j \neq i}^{m_2} \frac{1}{\left\|\overline{\boldsymbol{F}}_{i}-\overline{\boldsymbol{F}}_{j}\right\|^{2}}.
\end{equation}
This encourages fine-grained part discrimination within local regions and alleviates representation collapse among small object parts.

\begin{figure*}[t]
\centering
\includegraphics[width=1.0\linewidth]{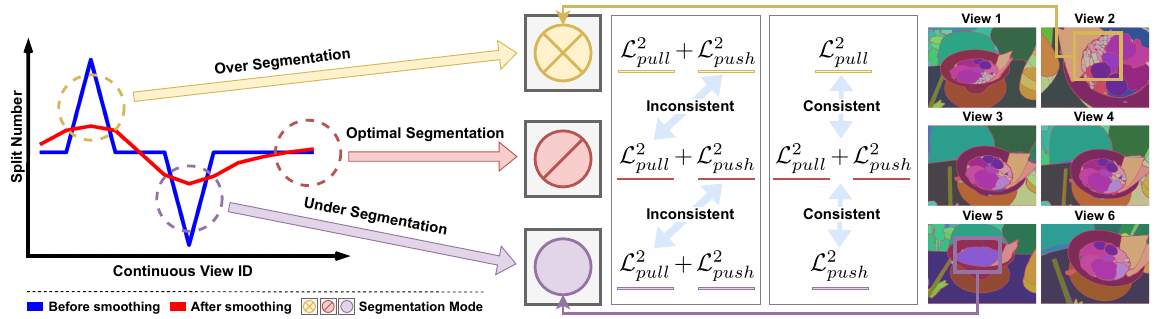} 
\caption{Consistent Segmentation Detection (CSD) for local contrastive learning. The blue curve shows the raw Split Number and the red curve shows the smoothed reference across views. Views where the blue curve lies above the red reference are treated as over segmentation (apply only $\mathcal{L}_{pull}^2$), while views where it lies below are treated as under segmentation (apply only $\mathcal{L}_{push}^2$). When the blue curve is close to the reference, we treat it as optimal segmentation (apply $\mathcal{L}_{pull}^2+\mathcal{L}_{push}^2$).}
\label{fig:consistency_detection}
\end{figure*}

\subsection{Consistent Segmentation Detection}
\label{sec:csd}

To prevent inconsistent supervision caused by multi-view object-tree mismatches, we introduce a Consistent Segmentation Detection (CSD) mechanism. It dynamically determines which contrastive component to apply at each view, enabling the model to better align the different mask segmentation modes. Specifically, for each continuous view $v$, we compute the overlap of the binary region $\boldsymbol{B}$ with the fine-scale masks via IoU. The number of matched fine masks at view $v$ defines the “Split Number” as $n_v$. As shown in Figure~\ref{fig:consistency_detection}, we smooth the raw Split Number $n_v$ (blue) with a sliding window across views, obtaining a smoothed reference $\bar{n}_v$ (red) for stable decision-making. Since $\bar{n}_v$ can be fractional, we quantize it to an integer reference $\hat{n}_v=\mathrm{round}(\bar{n}_v)$ and determine the segmentation mode by comparing $n_v$ with $\hat{n}_v$:
\begin{itemize}
  \item \textbf{Over Segmentation} ($n_v>\hat{n}_v$): apply only the pull loss $\mathcal{L}_{local}=\mathcal{L}_{pull}^2$ to merge over-fragmented clusters.
  \item \textbf{Under Segmentation} ($n_v<\hat{n}_v$): apply only the push loss $\mathcal{L}_{local}=\mathcal{L}_{push}^2$ to separate under-merged clusters.
  \item \textbf{Optimal Segmentation} ($n_v=\hat{n}_v$): apply the full local loss $\mathcal{L}_{local}=\mathcal{L}_{pull}^2 + \mathcal{L}_{push}^2$ to simultaneously refine and discriminate instance features, producing the most stable clusters.
\end{itemize}
This provides flexible cross-view dynamic supervision without forcing identical segmentation modes. Then we recursively perform localized training and clustering, hierarchically partitioning the scene into finer substructures until reaching a predefined maximum tree depth. The progressive learning process enables training stability and consistent feature representation in complex scenes.

\subsection{Graph-Based Gaussian Points Denoising}
\label{sec:denoising}

Despite the effectiveness of our tree-guided cascaded learning framework, we observe that noisy Gaussian points may still emerge in certain cases, which hinders the quality of the instance cluster. These noise points are typically introduced due to the following factors: 1) Erroneous ground-truth masks from SAM, especially in clutter or occlusion. 2) Lack of contrast for objects that never co-occur in the same view, causing feature collapse. 3) Inaccurate clustering that groups unrelated regions. Inspired by SUNDAE \citep{yang2024spectrally}, we employ a graph‐based denoising module to enhance cluster purity. For each instance cluster, we build a feature‐aware nearest neighbor graph over its Gaussian points, where edges connect points that are both spatially close and feature‐similar.  
Specifically, we define the adjacency matrix $W$, where each element $W_{ij}$ represents the edge weight between points $i$ and $j$, as follows:
\begin{equation}
\resizebox{\linewidth}{!}{
  $W_{ij} = 
  \left\{
  \begin{array}{cl}
  \exp\left(-\dfrac{\left\|p_i - p_j\right\|_2^2}{2\sigma_1^2}\right) + \exp\left(-\dfrac{\left\|\boldsymbol{f}_i - \boldsymbol{f}_j\right\|_2^2}{2\sigma_2^2}\right), &  \left\|p_i - p_j\right\|_2^2 < \tau_1 \text{ and } \left\|\boldsymbol{f}_i - \boldsymbol{f}_j\right\|_2^2 < \tau_2 \\
  0, & \text{otherwise}
  \end{array}
  \right.$
}
\end{equation}
where $p_i$ and $\boldsymbol{f}_i$ denote the position and instance feature of point $i$, and $\sigma_1, \sigma_2$ are the standard deviations computed from the pairwise distance distributions in the position and feature spaces. $\tau_1$ and $\tau_2$ are distance thresholds in the spatial and feature domains. The connection strength $C_i = \sum_j W_{ij}$ reflects the local density around the point $i$. We discard any point whose connection strength falls below the mean connection strength $\overline{C}$, treating it as likely noise. To prevent inadvertent removal of valid object points during the denoising process, we fit an oriented bounding box (OBB) \citep{gottschalk2000collision} around each denoised instance cluster and restore any points that were filtered out but lie within the box. This step ensures that non-noisy portions of the object are preserved for accurate multi‐view rendering. For 2D-3D language mapping, we follow the approach proposed in OpenGaussian \citep{wuopengaussian}, where a CLIP embedding is assigned to each multi-scale cluster to serve as a semantic anchor for open-vocabulary querying. Specifically, we aggregate the CLIP semantic features extracted from multi-view rendered images of each cluster and use the aggregated embedding as its semantic anchor. This multi-view aggregation yields a more stable and view-invariant semantic representation for language-based queries.

\section{Experiments}

\subsection{Open-Vocabulary 3D Object Selection}

\textbf{Settings.}
\textbf{1) Task:} Given an open-vocabulary text query, we extract its language feature using CLIP and compute the cosine similarity between the query feature and the language feature associated with each instance cluster. The most relevant Gaussian clusters are then selected for rendering to the 2D plane.
\textbf{2) Dataset and Metrics:} We evaluated the performance of our method on the Lerf\_ovs dataset \citep{qin2024langsplat}, which covers four scenes (figurines, teatime, ramen, and waldo\_kitchen) with annotated pixel-level semantic labels. We computed the mIoU and mAcc@0.25 scores by comparing the rendered 2D objects with the ground-truth 2D annotations.
\textbf{3) Baseline:} We compare our approach with recent methods based on 3D Gaussian Splatting (3DGS). Specifically, we include LangSplat \citep{qin2024langsplat} and LEGaussians \citep{LangEmbedGaussian}, which leverage language distillation to directly supervise the learning of semantic fields. In addition, we compare with OpenGaussians \citep{wuopengaussian} and InstanceGaussian \citep{li2024instancegaussian}, which utilize the flat contrastive learning method for segmentation of point-level instances, as well as the fused contrastive learning method, OmniSeg3D-GS \citep{ying2024omniseg3d}.

\setlength{\tabcolsep}{4pt}
\begin{table*}[t]
  \caption{Quantitative comparison of open-vocabulary 3D object selection on the Lerf\_ovs dataset. We report mIoU(\%) and mAcc@0.25(\%) scores.}
  \label{tab:lerf_combined}
  \centering
  \begin{tabular}{*{5}{c}}
    \toprule
    Scale & Methods & Type & mIoU(\%) & mAcc(\%) \\
    \midrule
    whole & LangSplat \citep{qin2024langsplat}              & Language distillation & 9.66  & 12.41 \\
    whole & LEGaussians \citep{LangEmbedGaussian}           & Language distillation & 16.21 & 23.82 \\
    whole & OmniSeg3D-GS \citep{ying2024omniseg3d}          & Fused contrastive learning & 38.55 & 63.29 \\
    whole & OpenGaussian \citep{wuopengaussian}             & Flat contrastive learning & 38.36 & 51.43 \\
    whole & InstanceGaussian \citep{li2024instancegaussian} & Flat contrastive learning & 45.30 & 58.44 \\
    whole & Ours                                            & Cascaded contrastive learning & \textbf{51.78} & \textbf{70.51} \\
    \midrule
    part  & OmniSeg3D-GS \citep{ying2024omniseg3d}          & Fused contrastive learning & 35.57 & 56.52 \\
    part  & OpenGaussian \citep{wuopengaussian}             & Flat contrastive learning & 39.16 & 57.01 \\
    part  & InstanceGaussian \citep{li2024instancegaussian} & Flat contrastive learning & 36.14 & 46.73 \\
    part  & Ours                                            & Cascaded contrastive learning & \textbf{44.10} & \textbf{60.44} \\
    \bottomrule
  \end{tabular}
\end{table*}

\begin{figure*}[t]
\centering
\includegraphics[width=1.0\linewidth]{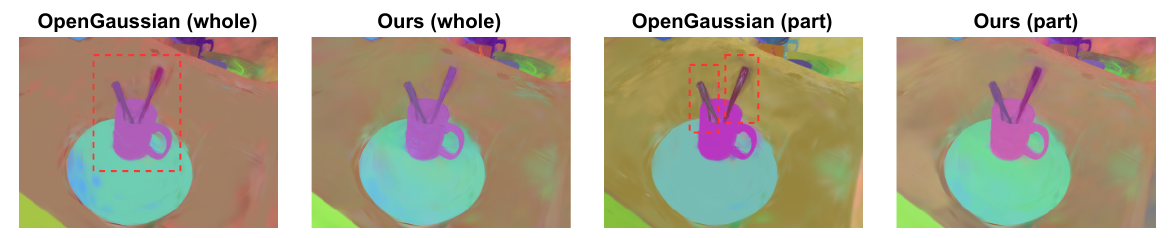}
\caption{Qualitative comparison of the rendered instance feature maps. Our method achieves better global feature consistency across objects (cup and spoon) at the whole scale and exhibits clearer feature separation at the part scale.}
\label{fig:lerf Qualitative comparison1}
\end{figure*}

\noindent
\textbf{Results.} The quantitative comparison are shown in Table~\ref{tab:lerf_combined}. Our two-stage cascaded contrastive learning strategy outperforms all baseline methods in both whole and part scale training results.
LangSplat and LEGaussians suffer from semantics loss caused by high-dimensional language feature compression, leading to poor performance in open-vocabulary object segmentation.
Flat contrastive learning methods, including OpenGaussians and InstanceGaussian, lack structural awareness. 
In complex scenes, hierarchical relationships between object wholes and their constituent parts play a crucial role in preserving semantic coherence and ensuring precise boundary delineation.
Without modeling hierarchical relationships, flat contrastive strategies are unable to produce accurate and coherent instance segmentation.
While OmniSeg3D-GS unifies whole and part feature spaces, the large number of redundant contrastive pairs introduces instability optimization results.
In contrast, our approach minimizes contrastive redundancy and explicitly addresses the issues of feature space saturation and multi-view hierarchical inconsistency, enabling robust instance segmentation and structured scene understanding across diverse and complex scenes.

\begin{figure*}[t]
\centering
\includegraphics[width=1.0\linewidth]{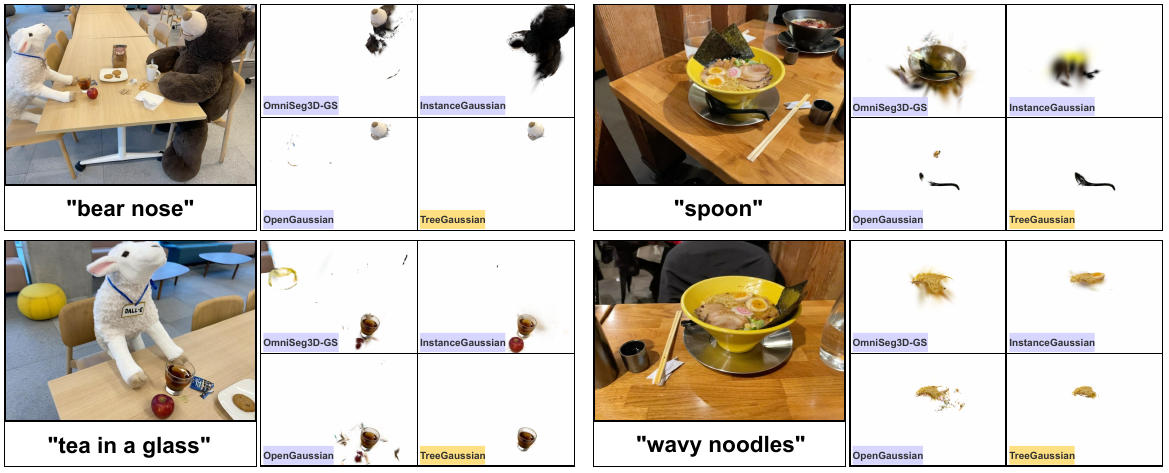}
\caption{Qualitative comparison of rendered local objects. Our method produces cleaner and more accurate segmentation results compared to baselines, effectively reducing noise and preserving fine-grained structures.}
\label{fig:Qualitative comparison2}
\end{figure*}

\begin{figure*}[t]
\centering
\includegraphics[width=1.0\linewidth]{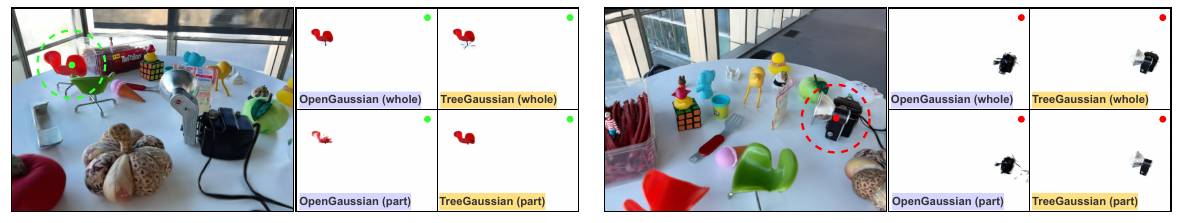}
\caption{Qualitative comparison of click-based 3D object selection. Compared to OpenGaussian, our method produces more accurate and hierarchically consistent results at both whole and part scales.}
\label{fig:exp_4}
\end{figure*}

The comparison of global instance feature maps is shown in Figure~\ref{fig:lerf Qualitative comparison1}. At the whole scale, OpenGaussian fails to optimize the features of the cup and the spoon into a coherent representation, while our method ensures global feature consistency. At the part scale, OpenGaussian struggles to separate the features of the cup and the spoon. Our method maintains a high degree of feature distinction between different parts. 
Figure~\ref{fig:Qualitative comparison2} presents various results on rendered local objects, illustrating the effectiveness of our method across different objects. Compared to baselines, our method yields more precise segmentation with less noise and better preservation of fine-grained structural details. To mitigate the semantic ambiguity introduced by CLIP language mapping, we additionally performed a click-based 3D object selection experiment. As shown in Figure~\ref{fig:exp_4}, our method yields more accurate results across different scales. Overall, our method achieves high-quality segmentation that not only ensures hierarchical consistency but also effectively suppresses noise and artifacts.

\setlength{\tabcolsep}{1.0pt}
\begin{table*}[ht]
\centering
\caption{Quantitative comparison on ScanNet-v2 dataset. We report mIoU(\%) scores.}
\label{tab:scannet_miou_all}
\begin{tabular}{c|ccc|ccc|ccc}
\toprule
\multirow{2}{*}{Methods}
& \multicolumn{3}{c|}{19 Classes}
& \multicolumn{3}{c|}{15 Classes}
& \multicolumn{3}{c}{10 Classes} \\
\cmidrule(lr){2-4} \cmidrule(lr){5-7} \cmidrule(lr){8-10}
& whole & part & subpart
& whole & part & subpart
& whole & part & subpart \\
\midrule
OmniSeg3D-GS \citep{ying2024omniseg3d}
& 20.56 & 15.29 & 12.11
& 23.74 & 17.83 & 18.79
& 32.41 & 25.73 & 18.79 \\

OpenGaussian \citep{wuopengaussian}
& 24.73 & 18.12 & 14.73
& 30.13 & 20.91 & 17.24
& 38.29 & 33.50 & 25.77 \\

InstanceGaussian \citep{li2024instancegaussian}
& 40.66 & 27.20 & 17.20
& 42.51 & 23.76 & 19.93
& 47.94 & 30.63 & 24.24 \\

Ours
& \textbf{41.61} & \textbf{28.99} & \textbf{18.32}
& \textbf{42.57} & \textbf{29.28} & \textbf{20.01}
& \textbf{49.80} & \textbf{38.76} & \textbf{27.66} \\

\bottomrule
\end{tabular}
\end{table*}

\setlength{\tabcolsep}{1.0pt}
\begin{table*}[ht]
\centering
\caption{Quantitative comparison on ScanNet-v2 dataset. We report mAcc(\%) scores.}
\label{tab:scannet_macc_all}
\begin{tabular}{c|ccc|ccc|ccc}
\toprule
\multirow{2}{*}{Methods}
& \multicolumn{3}{c|}{19 Classes}
& \multicolumn{3}{c|}{15 Classes}
& \multicolumn{3}{c}{10 Classes} \\
\cmidrule(lr){2-4} \cmidrule(lr){5-7} \cmidrule(lr){8-10}
& whole & part & subpart
& whole & part & subpart
& whole & part & subpart \\
\midrule
OmniSeg3D-GS \citep{ying2024omniseg3d}
& 38.51 & 25.39 & 23.13
& 36.02 & 29.06 & 30.49
& 48.32 & 38.91 & 30.49 \\

OpenGaussian \citep{wuopengaussian}
& 41.54 & 28.89 & 24.66
& 48.25 & 31.89 & 27.87
& 55.19 & 45.17 & 39.88 \\

InstanceGaussian  \citep{li2024instancegaussian}
& 54.01 & 40.73 & 27.78
& 59.15 & 37.63 & 29.07
& 64.01 & 48.89 & 37.50 \\

Ours
& \textbf{54.38} & \textbf{42.11} & \textbf{28.82}
& \textbf{59.94} & \textbf{42.95} & \textbf{32.61}
& \textbf{65.41} & \textbf{54.55} & \textbf{40.35} \\

\bottomrule
\end{tabular}
\end{table*}

\subsection{3D Point Cloud Understanding}

\textbf{Settings.}
\textbf{1) Task:}
Given a text query, the task aims to retrieve the corresponding point cloud by computing the cosine similarity between the text features and the point-level language features.
\textbf{2) Dataset and Metrics:}
Following the OpenGaussian evaluation protocol, we select 10 scenes from the ScanNet-v2 dataset \citep{dai2017scannet} and evaluate 19, 15 and 10 semantic categories as text queries. The ScanNet-v2 dataset provides posed RGB images, reconstructed point clouds, and 3D point-level semantic annotations. We initialize Gaussians from the provided point clouds, freezing coordinates and disabling densification during training to ensure alignment with the 3D point-level labels. For each query, we retrieve the most similar point cloud instance and compute the point-level mIoU and mAcc for each category as evaluation metrics. 
\textbf{3) Baseline:}
We compared with recent Gaussian-based contrastive learning approaches, including OmniSeg3D-GS, OpenGaussian, and InstanceGaussian.

\noindent
\textbf{Results.} The quantitative results are shown in Table~\ref{tab:scannet_miou_all} and Table~\ref{tab:scannet_macc_all}. Our method consistently outperforms the baselines across multiple scales (whole, part, and subpart). More substantial improvements are observed at finer scales (part and subpart), suggesting that our hierarchical design enables more effective supervision and sufficient learning of fine-grained local structures often overlooked by existing methods. However, as the scale decreases, the performance degrades more significantly. We attribute this to the difficulty of CLIP in extracting reliable language features for tiny objects, as such regions are often geometrically small and semantically ambiguous. Despite this limitation, our approach still achieves robust segmentation at all scales, underscoring the validity of our hierarchical semantic assumptions and cascaded contrastive learning strategy.

\subsection{Ablation Study}

\noindent
\textbf{(1) Effect of the Local Loss and Denoising Module.} Table~\ref{tab:ablation1} reports the ablation study that evaluates the impact of removing the local contrastive loss $\mathcal{L}_{local}$ and the graph-based denoising module. Excluding $\mathcal{L}_{local}$ results in a significant performance drop at both whole and part scales. For the whole scale, mIoU drops from 51.78 to 47.05 and mAcc from 70.51 to 62.92. For the part scale, performance degrades from 44.10 to 42.37 in mIoU and from 60.44 to 59.02 in mAcc. This demonstrates the critical role of local contrastive learning in enhancing feature discrimination. Removing the graph-based denoising module also leads to noticeable performance degradation, indicating its necessity to suppress noisy Gaussians and improve cluster purity. The full model with both techniques achieves the best performance on the Lerf\_ovs dataset. 

\setlength{\tabcolsep}{10pt}
\begin{table}[ht]
  \caption{Quantitative ablation results of Local Contrastive Loss and Graph-based Denoising on the Lerf\_ovs dataset. We report the
  mIoU(\%) and mAcc@0.25(\%) scores.}
  \label{tab:ablation1}
  \centering
  \begin{tabular}{*{7}{c}}  
    \toprule
    \multirow{2}{*}{Scale} & \multicolumn{2}{c}{w/o 
     $\mathcal{L}_{local}$} & \multicolumn{2}{c}{w/o Denoising} & \multicolumn{2}{c}{Ours} \\
    & mIoU & mAcc & mIoU & mAcc & mIoU & mAcc \\
    \midrule
    whole & 47.05 & 62.92 & 48.23 & 66.63 & \textbf{51.78} & \textbf{70.51} \\
    part  & 42.37 & 59.02 & 43.45 & 59.66 & \textbf{44.10} & \textbf{60.44} \\
    \bottomrule
  \end{tabular}
\end{table}

\noindent
\textbf{(2) Component-wise Analysis of Local Learning.}
To further investigate the contribution of each term in the local contrastive learning stage, we perform controlled ablations on the teatime scene as reported in Table \ref{tab:ablation2}. Removing $\mathcal{L}_{push}^2$ or the CSD strategy leads to a significant performance drop, especially for the whole scale, highlighting their critical role in inter-class separability and structural consistency. The absence of $\mathcal{L}_{pull}^2$ causes a moderate decline, suggesting its auxiliary role in feature compactness. For the part scale, all ablated variants exhibit limited performance. This suggests that without proper contrastive guidance, the model struggles to distinguish partially observed instances. Our full framework, by contrast, leverages the hierarchical local contrastive learning and the CSD mechanism to dynamically apply the appropriate loss. This allows the model to adapt to under-segmentation and over-segmentation modes by selectively applying $\mathcal{L}_{pull}^2$ or $\mathcal{L}_{push}^2$. As a result, it achieves significantly higher mAcc (74.58) and mIoU (53.68), indicating enhanced generalization to partial-object regions and more reliable semantic discrimination.

\setlength{\tabcolsep}{6pt}
\begin{table}[ht]
  \caption{Quantitative ablation results on the teatime scene, evaluating the effect of each local contrastive learning component.}
  \label{tab:ablation2}
  \centering
  \begin{tabular}{*{9}{c}}  
    \toprule
    \multirow{2}{*}{Scale} 
    & \multicolumn{2}{c}{w/o $\mathcal{L}_{pull}^2$} 
    & \multicolumn{2}{c}{w/o $\mathcal{L}_{push}^2$} 
    & \multicolumn{2}{c}{w/o CSD} 
    & \multicolumn{2}{c}{Ours} \\
    
    & mIoU & mAcc & mIoU & mAcc & mIoU & mAcc & mIoU & mAcc \\
    \midrule
    whole & 56.63 & 69.49 & 48.22 & 69.49 & 47.51 & 59.32 & \textbf{62.61} & \textbf{77.97} \\
    part  & 52.14 & 69.49 & 52.51 & 69.49 & 51.54 & 69.49 & \textbf{53.68} & \textbf{74.58} \\
    \bottomrule
  \end{tabular}
\end{table}

\begin{figure*}[t]
\centering
\includegraphics[width=1.0\linewidth]{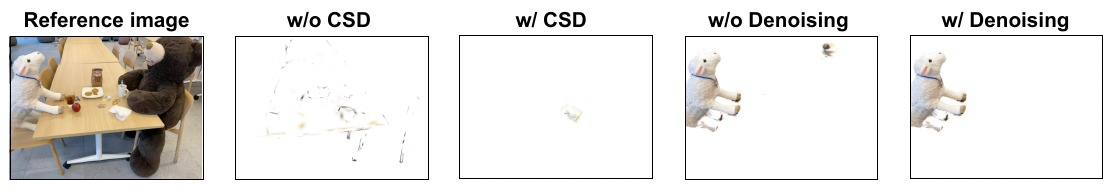}
\caption{Qualitative ablation results with Consistent Segmentation Detection (CSD) mechanism and Graph-based Gaussian Points Denoising. The results demonstrate that CSD reduces oversegmentation and enhances consistency, while the denoising module effectively suppresses distant clutter and improves clarity.}
\label{fig:ablation1}
\end{figure*}

\noindent
\textbf{(3) Qualitative Ablation.}
Figure~\ref{fig:ablation1} provides qualitative comparisons on the teatime scene. Without the CSD strategy, the proposed approach tends to oversegment the scene, resulting in fragmented and inconsistent outputs. Introducing CSD significantly alleviates these artifacts and improves overall spatial coherence. In the absence of graph-based denoising, spurious Gaussian clusters appear far from the target regions due to erroneous features or inaccurate clustering. Applying the denoising strategy effectively suppresses such outliers, producing cleaner and more spatially consistent instance representations. These visual results are highly consistent with the quantitative findings and further support the overall effectiveness of the proposed approach.

\section{Conclusion}

In this paper, we propose TreeGaussian, a 3DGS-based framework for open-vocabulary 3D point-level semantic segmentation and scene understanding. Existing methods often overlook the hierarchical structure of objects and suffer from low-dimensional feature saturation due to dense pairwise contrastive learning. To address these issues, we introduce a tree-guided method that models multi-level semantic relationships and reduces contrastive redundancy. We design a cascaded contrastive learning strategy to progressively refine instance-level representations and enhance training stability. In addition, we introduce a targeted consistency-aware segmentation mechanism to address the common issue of multi-view inconsistency, aligning segmentation mode across viewpoints for more stable and reliable predictions. A graph-based denoising module further improves the quality of instance clusters. Overall, TreeGaussian highlights the importance of structured hierarchical modeling in 3D scene understanding and provides more consistent multi-view scene reasoning in 3DGS representations.

\noindent
\textbf{Limitations: } (1) While CLIP provides strong semantic priors, it often struggles to represent the fine-grained semantics of tiny object parts. Leveraging hierarchical representations with more advanced vision-language models \citep{oquab2023dinov2, li2023blip} has the potential to further improve scene understanding. (2) Our method is currently limited to static scenes. Extending TreeGaussian to dynamic environments with temporal consistency represents a promising future direction. 




%
%

\bibliographystyle{splncs04}
\bibliography{main}

@String(AAAI  = {AAAI})

@inproceedings{li2023blip,
  title={Blip-2: Bootstrapping language-image pre-training with frozen image encoders and large language models},
  author={Li, Junnan and Li, Dongxu and Savarese, Silvio and Hoi, Steven},
  booktitle={International conference on machine learning},
  pages={19730--19742},
  year={2023},
  organization={PMLR}
}

@article{liu2019point,
  title={Point-voxel cnn for efficient 3d deep learning},
  author={Liu, Zhijian and Tang, Haotian and Lin, Yujun and Han, Song},
  journal={Advances in neural information processing systems},
  volume={32},
  year={2019}
}

@inproceedings{han2020occuseg,
  title={Occuseg: Occupancy-aware 3d instance segmentation},
  author={Han, Lei and Zheng, Tian and Xu, Lan and Fang, Lu},
  booktitle={Proceedings of the IEEE/CVF conference on computer vision and pattern recognition},
  pages={2940--2949},
  year={2020}
}

@inproceedings{yi2019gspn,
  title={Gspn: Generative shape proposal network for 3d instance segmentation in point cloud},
  author={Yi, Li and Zhao, Wang and Wang, He and Sung, Minhyuk and Guibas, Leonidas J},
  booktitle={Proceedings of the IEEE/CVF conference on computer vision and pattern recognition},
  pages={3947--3956},
  year={2019}
}

@article{yang2019learning,
  title={Learning object bounding boxes for 3d instance segmentation on point clouds},
  author={Yang, Bo and Wang, Jianan and Clark, Ronald and Hu, Qingyong and Wang, Sen and Markham, Andrew and Trigoni, Niki},
  journal={Advances in neural information processing systems},
  volume={32},
  year={2019}
}

@inproceedings{caron2021emerging,
  title={Emerging properties in self-supervised vision transformers},
  author={Caron, Mathilde and Touvron, Hugo and Misra, Ishan and J{\'e}gou, Herv{\'e} and Mairal, Julien and Bojanowski, Piotr and Joulin, Armand},
  booktitle={Proceedings of the IEEE/CVF international conference on computer vision},
  pages={9650--9660},
  year={2021}
}

@article{cen2023segment,
  title={Segment anything in 3d with nerfs},
  author={Cen, Jiazhong and Zhou, Zanwei and Fang, Jiemin and Shen, Wei and Xie, Lingxi and Jiang, Dongsheng and Zhang, Xiaopeng and Tian, Qi and others},
  journal={Advances in Neural Information Processing Systems},
  volume={36},
  pages={25971--25990},
  year={2023}
}

@inproceedings{aliev2020neural,
  title={Neural point-based graphics},
  author={Aliev, Kara-Ali and Sevastopolsky, Artem and Kolos, Maria and Ulyanov, Dmitry and Lempitsky, Victor},
  booktitle={European conference on computer vision},
  pages={696--712},
  year={2020},
  organization={Springer}
}

@inproceedings{schult2020dualconvmesh,
  title={Dualconvmesh-net: Joint geodesic and euclidean convolutions on 3d meshes},
  author={Schult, Jonas and Engelmann, Francis and Kontogianni, Theodora and Leibe, Bastian},
  booktitle={Proceedings of the IEEE/CVF conference on computer vision and pattern recognition},
  pages={8612--8622},
  year={2020}
}

@inproceedings{valentin2013mesh,
  title={Mesh based semantic modelling for indoor and outdoor scenes},
  author={Valentin, Julien PC and Sengupta, Sunando and Warrell, Jonathan and Shahrokni, Ali and Torr, Philip HS},
  booktitle={Proceedings of the IEEE Conference on Computer Vision and Pattern Recognition},
  pages={2067--2074},
  year={2013}
}

@inproceedings{peng2023openscene,
  title={Openscene: 3d scene understanding with open vocabularies},
  author={Peng, Songyou and Genova, Kyle and Jiang, Chiyu and Tagliasacchi, Andrea and Pollefeys, Marc and Funkhouser, Thomas and others},
  booktitle={Proceedings of the IEEE/CVF conference on computer vision and pattern recognition},
  pages={815--824},
  year={2023}
}

@inproceedings{fridovich2022plenoxels,
  title={Plenoxels: Radiance fields without neural networks},
  author={Fridovich-Keil, Sara and Yu, Alex and Tancik, Matthew and Chen, Qinhong and Recht, Benjamin and Kanazawa, Angjoo},
  booktitle={Proceedings of the IEEE/CVF conference on computer vision and pattern recognition},
  pages={5501--5510},
  year={2022}
}

@article{kerbl20233d,
  title={3d gaussian splatting for real-time radiance field rendering.},
  author={Kerbl, Bernhard and Kopanas, Georgios and Leimk{\"u}hler, Thomas and Drettakis, George},
  journal={ACM Trans. Graph.},
  volume={42},
  number={4},
  pages={139--1},
  year={2023}
}

@article{mildenhall2021nerf,
  title={Nerf: Representing scenes as neural radiance fields for view synthesis},
  author={Mildenhall, Ben and Srinivasan, Pratul P and Tancik, Matthew and Barron, Jonathan T and Ramamoorthi, Ravi and Ng, Ren},
  journal={Communications of the ACM},
  volume={65},
  number={1},
  pages={99--106},
  year={2021},
  publisher={ACM New York, NY, USA}
}

@article{yang2024spectrally,
  title={Spectrally pruned gaussian fields with neural compensation},
  author={Yang, Runyi and Zhu, Zhenxin and Jiang, Zhou and Ye, Baijun and Chen, Xiaoxue and Zhang, Yifei and Chen, Yuantao and Zhao, Jian and Zhao, Hao},
  journal={arXiv preprint arXiv:2405.00676},
  year={2024}
}

@inproceedings{qin2024langsplat,
  title={Langsplat: 3d language gaussian splatting},
  author={Qin, Minghan and Li, Wanhua and Zhou, Jiawei and Wang, Haoqian and Pfister, Hanspeter},
  booktitle={Proceedings of the IEEE/CVF Conference on Computer Vision and Pattern Recognition},
  pages={20051--20060},
  year={2024}
}

@inproceedings{zhou2024feature,
  title={Feature 3dgs: Supercharging 3d gaussian splatting to enable distilled feature fields},
  author={Zhou, Shijie and Chang, Haoran and Jiang, Sicheng and Fan, Zhiwen and Zhu, Zehao and Xu, Dejia and Chari, Pradyumna and You, Suya and Wang, Zhangyang and Kadambi, Achuta},
  booktitle={Proceedings of the IEEE/CVF Conference on Computer Vision and Pattern Recognition},
  pages={21676--21685},
  year={2024}
}

@inproceedings{radford2021learning,
  title={Learning transferable visual models from natural language supervision},
  author={Radford, Alec and Kim, Jong Wook and Hallacy, Chris and Ramesh, Aditya and Goh, Gabriel and Agarwal, Sandhini and Sastry, Girish and Askell, Amanda and Mishkin, Pamela and Clark, Jack and others},
  booktitle={International conference on machine learning},
  pages={8748--8763},
  year={2021},
  organization={PmLR}
}

@inproceedings{wuopengaussian,
 author = {Wu, Yanmin and Meng, Jiarui and Li, Haijie and Wu, Chenming and Shi, Yahao and Cheng, Xinhua and Zhao, Chen and Feng, Haocheng and Ding, Errui and Wang, Jingdong and Zhang, Jian},
 booktitle = {Advances in Neural Information Processing Systems},
 editor = {A. Globerson and L. Mackey and D. Belgrave and A. Fan and U. Paquet and J. Tomczak and C. Zhang},
 pages = {19114--19138},
 publisher = {Curran Associates, Inc.},
 title = {OpenGaussian: Towards Point-Level 3D Gaussian-based Open Vocabulary Understanding},
 url = {https://proceedings.neurips.cc/paper_files/paper/2024/file/21f7b745f73ce0d1f9bcea7f40b1388e-Paper-Conference.pdf},
 volume = {37},
 year = {2024}
}

@inproceedings{li2024instancegaussian,
  title={Instancegaussian: Appearance-semantic joint gaussian representation for 3d instance-level perception},
  author={Li, Haijie and Wu, Yanmin and Meng, Jiarui and Gao, Qiankun and Zhang, Zhiyao and Wang, Ronggang and Zhang, Jian},
  booktitle={Proceedings of the Computer Vision and Pattern Recognition Conference},
  pages={14078--14088},
  year={2025}
}

@inproceedings{choi2024click,
  title={Click-gaussian: Interactive segmentation to any 3d gaussians},
  author={Choi, Seokhun and Song, Hyeonseop and Kim, Jaechul and Kim, Taehyeong and Do, Hoseok},
  booktitle={European Conference on Computer Vision},
  pages={289--305},
  year={2024},
  organization={Springer}
}

@inproceedings{cen2025segment,
  title={Segment any 3d gaussians},
  author={Cen, Jiazhong and Fang, Jiemin and Yang, Chen and Xie, Lingxi and Zhang, Xiaopeng and Shen, Wei and Tian, Qi},
  booktitle={Proceedings of the AAAI Conference on Artificial Intelligence},
  pages={1971--1979},
  year={2025}
}

@inproceedings{segmentAnything,
  title={Segment anything},
  author={Kirillov, Alexander and Mintun, Eric and Ravi, Nikhila and Mao, Hanzi and Rolland, Chloe and Gustafson, Laura and Xiao, Tete and Whitehead, Spencer and Berg, Alexander C and Lo, Wan-Yen and others},
  booktitle={Proceedings of the IEEE/CVF international conference on computer vision},
  pages={4015--4026},
  year={2023}
}

@inproceedings{n2f2,
  title={N2f2: Hierarchical scene understanding with nested neural feature fields},
  author={Bhalgat, Yash and Laina, Iro and Henriques, Jo{\~a}o F and Zisserman, Andrew and Vedaldi, Andrea},
  booktitle={European Conference on Computer Vision},
  pages={197--214},
  year={2024},
  organization={Springer}
}

@inproceedings{kerr2023lerf,
  title={Lerf: Language embedded radiance fields},
  author={Kerr, Justin and Kim, Chung Min and Goldberg, Ken and Kanazawa, Angjoo and Tancik, Matthew},
  booktitle={Proceedings of the IEEE/CVF International Conference on Computer Vision},
  pages={19729--19739},
  year={2023}
}

@inproceedings{ying2024omniseg3d,
  title={Omniseg3d: Omniversal 3d segmentation via hierarchical contrastive learning},
  author={Ying, Haiyang and Yin, Yixuan and Zhang, Jinzhi and Wang, Fan and Yu, Tao and Huang, Ruqi and Fang, Lu},
  booktitle={Proceedings of the IEEE/CVF Conference on Computer Vision and Pattern Recognition},
  pages={20612--20622},
  year={2024}
}

@article{li2022lseg,
  title={Language-driven semantic segmentation},
  author={Li, Boyi and Weinberger, Kilian Q and Belongie, Serge and Koltun, Vladlen and Ranftl, Ren{\'e}},
  journal={arXiv preprint arXiv:2201.03546},
  year={2022}
}

@inproceedings{caron2021dino,
  title={Emerging properties in self-supervised vision transformers},
  author={Caron, Mathilde and Touvron, Hugo and Misra, Ishan and J{\'e}gou, Herv{\'e} and Mairal, Julien and Bojanowski, Piotr and Joulin, Armand},
  booktitle={Proceedings of the IEEE/CVF international conference on computer vision},
  pages={9650--9660},
  year={2021}
}

@book{gottschalk2000collision,
  title={Collision queries using oriented bounding boxes},
  author={Gottschalk, Stefan Aric},
  year={2000},
  publisher={The University of North Carolina at Chapel Hill}
}

@inproceedings{LangEmbedGaussian,
  title={Language embedded 3d gaussians for open-vocabulary scene understanding},
  author={Shi, Jin-Chuan and Wang, Miao and Duan, Hao-Bin and Guan, Shao-Hua},
  booktitle={Proceedings of the IEEE/CVF Conference on Computer Vision and Pattern Recognition},
  pages={5333--5343},
  year={2024}
}

@inproceedings{dai2017scannet,
  title={Scannet: Richly-annotated 3d reconstructions of indoor scenes},
  author={Dai, Angela and Chang, Angel X and Savva, Manolis and Halber, Maciej and Funkhouser, Thomas and Nie{\ss}ner, Matthias},
  booktitle={Proceedings of the IEEE conference on computer vision and pattern recognition},
  pages={5828--5839},
  year={2017}
}

@inproceedings{kim2024garfield,
  title={Garfield: Group anything with radiance fields},
  author={Kim, Chung Min and Wu, Mingxuan and Kerr, Justin and Goldberg, Ken and Tancik, Matthew and Kanazawa, Angjoo},
  booktitle={Proceedings of the IEEE/CVF Conference on Computer Vision and Pattern Recognition},
  pages={21530--21539},
  year={2024}
}

@inproceedings{lassner2021neuralRendering,
  title={Pulsar: Efficient sphere-based neural rendering},
  author={Lassner, Christoph and Zollhofer, Michael},
  booktitle={Proceedings of the IEEE/CVF Conference on Computer Vision and Pattern Recognition},
  pages={1440--1449},
  year={2021}
}

@book{foley1996alphaBlending,
  title={Computer graphics: principles and practice},
  author={Foley, James D},
  volume={12110},
  year={1996},
  publisher={Addison-Wesley Professional}
}

@inproceedings{fridovich2022sphericalHarmonics,
  title={Plenoxels: Radiance fields without neural networks},
  author={Fridovich-Keil, Sara and Yu, Alex and Tancik, Matthew and Chen, Qinhong and Recht, Benjamin and Kanazawa, Angjoo},
  booktitle={Proceedings of the IEEE/CVF conference on computer vision and pattern recognition},
  pages={5501--5510},
  year={2022}
}

@article{kusupati2022nestedfeature,
  title={Matryoshka representation learning},
  author={Kusupati, Aditya and Bhatt, Gantavya and Rege, Aniket and Wallingford, Matthew and Sinha, Aditya and Ramanujan, Vivek and Howard-Snyder, William and Chen, Kaifeng and Kakade, Sham and Jain, Prateek and others},
  journal={Advances in Neural Information Processing Systems},
  volume={35},
  pages={30233--30249},
  year={2022}
}

@inproceedings{wang2020understanding,
  title={Understanding contrastive representation learning through alignment and uniformity on the hypersphere},
  author={Wang, Tongzhou and Isola, Phillip},
  booktitle={International conference on machine learning},
  pages={9929--9939},
  year={2020},
  organization={PMLR}
}

@inproceedings{guadarrama2014open,
  title={Open-vocabulary Object Retrieval.},
  author={Guadarrama, Sergio and Rodner, Erik and Saenko, Kate and Zhang, Ning and Farrell, Ryan and Donahue, Jeff and Darrell, Trevor},
  booktitle={Robotics: science and systems},
  volume={2},
  pages={6},
  year={2014}
}

@inproceedings{mccormac2017semanticfusion,
  title={Semanticfusion: Dense 3d semantic mapping with convolutional neural networks},
  author={McCormac, John and Handa, Ankur and Davison, Andrew and Leutenegger, Stefan},
  booktitle={2017 IEEE International Conference on Robotics and automation (ICRA)},
  pages={4628--4635},
  year={2017},
  organization={IEEE}
}

@article{mur2017orb,
  title={Orb-slam2: An open-source slam system for monocular, stereo, and rgb-d cameras},
  author={Mur-Artal, Raul and Tard{\'o}s, Juan D},
  journal={IEEE transactions on robotics},
  volume={33},
  number={5},
  pages={1255--1262},
  year={2017},
  publisher={IEEE}
}

@article{humblot2022navigation,
  title={Navigation-oriented scene understanding for robotic autonomy: Learning to segment driveability in egocentric images},
  author={Humblot-Renaux, Galadrielle and Marchegiani, Letizia and Moeslund, Thomas B and Gade, Rikke},
  journal={IEEE Robotics and Automation Letters},
  volume={7},
  number={2},
  pages={2913--2920},
  year={2022},
  publisher={IEEE}
}

@article{saxena2008robotic,
  title={Robotic grasping of novel objects using vision},
  author={Saxena, Ashutosh and Driemeyer, Justin and Ng, Andrew Y},
  journal={The International Journal of Robotics Research},
  volume={27},
  number={2},
  pages={157--173},
  year={2008},
  publisher={Sage Publications Sage UK: London, England}
}

@article{zheng2024gaussiangrasper,
  title={Gaussiangrasper: 3d language gaussian splatting for open-vocabulary robotic grasping},
  author={Zheng, Yuhang and Chen, Xiangyu and Zheng, Yupeng and Gu, Songen and Yang, Runyi and Jin, Bu and Li, Pengfei and Zhong, Chengliang and Wang, Zengmao and Liu, Lina and others},
  journal={IEEE Robotics and Automation Letters},
  year={2024},
  publisher={IEEE}
}

@inproceedings{li2024object,
  title={Object-aware gaussian splatting for robotic manipulation},
  author={Li, Yulong and Pathak, Deepak},
  booktitle={ICRA 2024 Workshop on 3D Visual Representations for Robot Manipulation},
  year={2024}
}

@inproceedings{macqueen1967kmeans,
  title={Some methods for classification and analysis of multivariate observations},
  author={MacQueen, James},
  booktitle={Proceedings of the Fifth Berkeley Symposium on Mathematical Statistics and Probability, Volume 1: Statistics},
  volume={5},
  pages={281--298},
  year={1967},
  organization={University of California press}
}

@article{oquab2023dinov2,
  title={Dinov2: Learning robust visual features without supervision},
  author={Oquab, Maxime and Darcet, Timoth{\'e}e and Moutakanni, Th{\'e}o and Vo, Huy and Szafraniec, Marc and Khalidov, Vasil and Fernandez, Pierre and Haziza, Daniel and Massa, Francisco and El-Nouby, Alaaeldin and others},
  journal={arXiv preprint arXiv:2304.07193},
  year={2023}
}

\clearpage

\begin{center}
{\Large\bfseries Supplementary Material\par}
\end{center}

\setcounter{page}{1}

\setcounter{figure}{0}
\renewcommand{\thefigure}{S\arabic{figure}}
\setcounter{table}{0}
\renewcommand{\thetable}{S\arabic{table}}
\setcounter{section}{0} 
\renewcommand{\thesection}{\Alph{section}} 

\section{Implementation Details}

\subsection{Object‑tree Construction Details}

\begin{figure*}[h]
\centering
\includegraphics[width=1.0\linewidth]{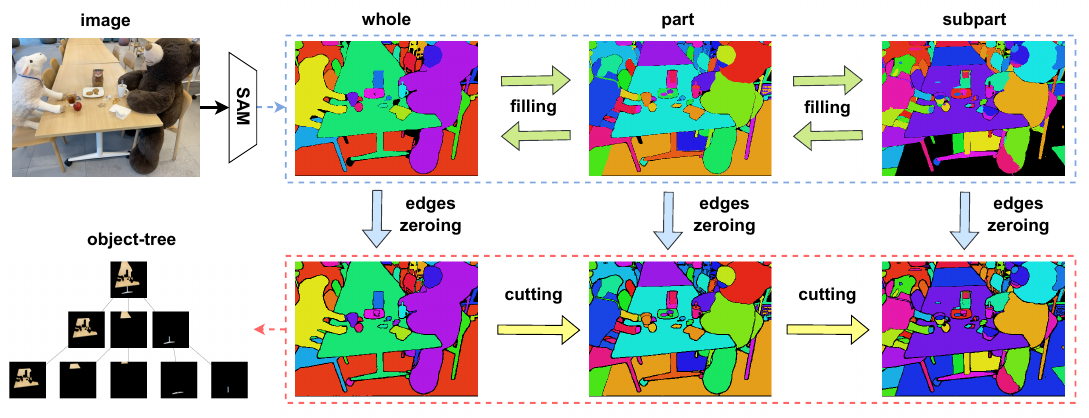}
\caption{More detailed visualization of multi-level object-tree construction}
\label{fig:Appendices1}
\end{figure*}

Figure~\ref{fig:Appendices1} illustrates the construction of a multi-level object hierarchy (whole, part, and subpart) from the initial SAM segmentation. At each level, the masks are refined through filling and edge-zeroing to adjust the granularity, followed by a cutting step for mask extraction. This supplementary figure provides an additional visualization of the object-tree construction process used in our framework. For clarity and dataset consistency, we present a three-level hierarchy in this example. However, the proposed cascaded contrastive learning framework is not inherently restricted to this specific hierarchy depth. Additional semantic levels can be incorporated by recursively extending the cascade depth, with each level operating on progressively refined instance features without requiring architectural changes.

\subsection{Training Strategy}

Consistent with OpenGaussian, we first pre-train the standard 3DGS model for 30,000 steps. Then, freezing all Gaussian parameters except the instance feature, we fine-tune the instance feature under different mask scales. 
On the Lerf\_ovs dataset, we trained for 50,000 steps at the whole scale and 100,000 steps at the part scale. On the ScanNet-v2 dataset, we trained for 60,000 steps at both the whole and part scales, and 80,000 steps at the subpart scale. All experiments were conducted on a single NVIDIA A100 GPU. 

\subsection{Hyperparameters}

(1) For all datasets, we set the instance feature dimensionality $D = 6$. 
(2) During multi-view 2D object-tree construction, masks smaller than 2,500 pixels are discarded.
(3) In CSD, we apply a sliding window of size 9 across views to smooth the raw Split Number curve.
(4) Graph-based Denoising: the spatial threshold $\tau_1$ is set to 100$\times$ the minimum non-zero distance between point positions, and the feature threshold $\tau_2$ is set to 50$\times$ the minimum non-zero distance between instance features. These values ensure that connections are made only between nearby and semantically similar points. 
The size of the OBB box is set to 1.2$\times$ the tightly fitting size.

\subsection{Evaluation Details}

Our method is evaluated at multiple semantic scales. Table 1 in the main paper reports results at the whole and part scales, while Table 2 further includes subpart-scale evaluation. These results are obtained using SAM-generated masks as supervision signals during training to optimize the instance features of 3D Gaussians. 

For OpenGaussian and InstanceGaussian evaluation, both methods rely on SAM-generated multi-level masks. The official implementations only report evaluation results at the whole scale using SAM masks. This does not fully reflect performance differences under finer-grained semantic hierarchies. We further extend the evaluation to finer scales, including part-scale results on the Lerf\_ovs dataset and part- and subpart-scale results on the ScanNet-v2 dataset. 

For OmniSeg3D evaluation, we follow the official instance retrieval protocol by sweeping a cosine-similarity threshold over the per-point instance features. OmniSeg3D originally supports only click-based querying. To enable open-vocabulary language evaluation, we introduce an additional language-to-click conversion stage that transforms language queries into click queries compatible with OmniSeg3D without modifying the original retrieval mechanism. Specifically, we generate multiple candidate query regions from multi-level masks on the test views. CLIP is used to measure the similarity between the language query and each candidate region, and the region with the highest similarity is selected to produce a click query point. The selected click is encoded into an instance feature and matched against the 3D feature field, strictly following the original OmniSeg3D inference pipeline. To evaluate multi-scale performance, cosine-similarity thresholds are grouped into three ranges on the ScanNet-v2 dataset for whole, part, and subpart scales: $[0.0,0.75]$, $[0.75,0.95]$, and $[0.95,1.0]$, and into two ranges on the Lerf\_ovs dataset for whole and part scales: $[0.0,0.75]$ and $[0.75,1.0]$. We sample 20 thresholds per range and report the best result at each scale.

\subsection{Dataset Details}

\noindent
\textbf{Lerf\_ovs dataset.} 
We evaluate our method on the Lerf\_ovs dataset, which consists of 4 scenes: figurines, teatime, ramen, and waldo\_kitchen.  Each scene is annotated with pixel-level text labels, enabling a fine-grained evaluation of semantic query. The annotated object categories are listed in Table~\ref{tab:lerf_ovs_annotated}. These detailed annotations enable evaluation of the method's performance across different object types and environments.

\noindent
\textbf{ScanNet-v2 dataset.} 
Following OpenGaussian, we select 10 scenes from the ScanNet-v2 dataset for evaluation: scene0000\_00, scene0062\_00, scene0070\_00, scene0097\_00, scene0140\_00, scene0200\_00, scene0347\_00, scene0400\_00, 

\noindent
scene0590\_00 and scene0645\_00. For text-based queries, we adopt 19, 15 and 10 semantic categories, as summarized in Table~\ref{tab:scannet_evaluation_Annotated}. All scene selections and category definitions follow the protocol used in OpenGaussian to ensure consistent comparison with existing baselines. 

\begin{table}[h]
  \caption{Annotated object categories for each scene in the Lerf\_ovs dataset.}
  \label{tab:lerf_ovs_annotated}
  \centering
  \begin{tabularx}{0.95\linewidth}{lX}
    \toprule
    \textbf{Scene} & \multicolumn{1}{c}{\textbf{Annotated Object Categories}} \\
    \midrule
    \textbf{figurines} & \textit{jake, pirate hat, pikachu, rubber duck with hat, porcelain hand, red apple, tesla door handle, waldo, bag, toy cat statue, miffy, green apple, pumpkin, rubics cube, old camera, rubber duck with buoy, red toy chair, pink ice cream, spatula, green toy chair}, and \textit{toy elephant}. \\
    \midrule
    \textbf{teatime} & \textit{sheep, yellow pouf, stuffed bear, coffee mug, tea in a glass, apple, coffee, hooves, bear nose, dall-e brand, plate, paper napkin, three cookies}, and \textit{bag of cookies}. \\
    \midrule
    \textbf{ramen} & \textit{nori, sake cup, kamaboko, corn, spoon, egg, onion segments, plate, napkin, bowl, glass of water, hand, chopsticks}, and \textit{wavy noodles}. \\
    \midrule
    \textbf{waldo\_kitchen} & \textit{stainless steel pots, dark cup, refrigerator, frog cup, pot, spatula, plate, spoon, toaster, ottolenghi, plastic ladle, sink, ketchup, cabinet, red cup, pour-over vessel, knife}, and \textit{yellow desk}. \\
    \bottomrule
  \end{tabularx}
\end{table}

\begin{table}[h]
  \caption{Annotated semantic categories for 10 scenes in the ScanNet-v2 dataset.}
  \label{tab:scannet_evaluation_Annotated}
  \centering
  \begin{tabular}{p{0.2\linewidth} p{0.65\linewidth}} 
    \toprule
    \textbf{19 Classes} & \textit{wall, floor, cabinet, bed, chair, sofa, table, door, window, bookshelf, picture, counter, desk, curtain, refrigerator, shower curtain, toilet, sink, bathtub} \\
    \midrule
    \textbf{15 Classes} & \textit{wall, floor, cabinet, bed, chair, sofa, table, door, window, bookshelf, counter, desk, toilet, sink, bathtub} \\
    \midrule
    \textbf{10 Classes} & \textit{wall, floor, bed, chair, sofa, table, door, window, toilet, sink} \\
    \bottomrule
  \end{tabular}
\end{table}


\clearpage

\section{More Results}

To further demonstrate the effectiveness of our method, We provide more experimental and visualization results.

\subsection{Additional Methods Comparison}

In the main paper, we compare flat semantic-assumption methods that support multi-level whole/part querying (OpenGaussian and InstanceGaussian), as well as a fused multi-level method (OmniSeg3D). Here we further include an additional NeRF-based fused method (GarField) on the Lerf\_ovs dataset. GarField is evaluated using the same cosine-similarity threshold settings as OmniSeg3D, where thresholds are grouped into two ranges on the Lerf\_ovs dataset for whole and part scales: $[0.0,0.75]$ and $[0.75,1.0]$, with 20 thresholds uniformly sampled in each range. For each scale, we report the best result within the corresponding threshold range. As shown in Table~\ref{tab:lerf_ovs_garfield}, our 3DGS-based tree-guided framework consistently outperforms this NeRF-based baseline on both whole and part scales, indicating that the proposed hierarchical supervision and cascaded optimization are effective beyond the underlying scene representation.

\begin{table}[h]
  \centering
  \caption{Comparison with an additional NeRF-based method on Lerf\_ovs dataset.}
  \label{tab:lerf_ovs_garfield}
  \resizebox{\linewidth}{!}{%
  \begin{tabular}{c c c c c c}
    \toprule
    Methods & Type & mIoU (whole) & mAcc (whole) & mIoU (part)  & mAcc (part) \\
    \midrule
    GarField & Fused & 49.91 & 65.62 & 40.15 & 59.14 \\
    Ours     & Cascaded & \textbf{51.78}  & \textbf{70.51} & \textbf{44.10} & \textbf{60.44} \\
    \bottomrule
  \end{tabular}
  }
\end{table}

\subsection{Runtime and Memory Analysis}

We provide a detailed computational cost breakdown of our method, focusing on the efficiency and scalability of our pipeline. The analysis includes the SAM preprocessing time per image, the training time per-scale, and the VRAM usage. We also report the additional computational overhead introduced by CSD and denoising techniques. This breakdown allows us to evaluate the efficiency of our approach and highlight areas where the computational load is most concentrated. The results, summarized in Table~\ref{tab:cost_breakdown}, show that for both datasets (Lerf\_ovs and ScanNet-v2). SAM preprocessing time with Lerf\_ovs requiring slightly more time per image due to its higher resolution. Training time and VRAM are linearly correlated with the complexity of the task. The whole-scale model requires the least time, and the subpart scale takes the longest. The overhead introduced by CSD remains minimal, contributing less than 1 minute to the overall computation time. In contrast, denoising adds a modest but important overhead, with processing times around 1.2 seconds per cluster for Lerf\_ovs and 0.8 seconds per cluster for ScanNet-v2. This denoising step is crucial to improve segmentation quality, especially in noisy scenes.

\begin{table}[h]
\centering
\renewcommand{\arraystretch}{1.1}
\caption{Computational cost breakdown of our pipeline on Lerf\_ovs and ScanNet-v2. We report SAM runtime per image, training time and VRAM at different semantic scales, and the overhead of CSD and denoising.}
\label{tab:cost_breakdown}
\resizebox{\linewidth}{!}{%
\begin{tabular}{ccccccc}
\toprule
Dataset & SAM (s/img) & Scale & Training time (min) & VRAM (GB) & CSD & Denoising  \\
\midrule
Lerf\_ovs  & 26.85 & whole / part & 30 / 50  & 7.6 / 10.9 & $<$ 1.0 min & 1.2 s/cluster \\
ScanNet-v2 & 19.15 & whole / part / subpart & 20 / 30 / 30 & 9.9 / 14.4 / 16.8 & $<$ 1.0 min & 0.8 s/cluster \\
\bottomrule
\end{tabular}
}
\end{table}

\subsection{Additional Ablation Results}

Table~\ref{tab:ablation_on_ScanNet} provides additional ablation results for the 3D point cloud understanding task, specifically focusing on the ScanNet-v2 dataset with the part scale. In this evaluation, we report results across 19 distinct classes, providing additional support for the model's generalization. The results clearly highlight the contribution of each individual component to the overall performance of the model. Among the tested configurations, the full model consistently outperforms all others, achieving the highest scores in both point cloud mean Intersection over Union (mIoU) and mean Accuracy (mAcc). This demonstrates the importance of each component, underscoring their collective impact on enhancing the model's ability to accurately segment 3D point cloud data.

\begin{table}[h]
    \centering
    \caption{Additional ablation results on the ScanNet-v2 dataset with part scale.}
    \label{tab:ablation_on_ScanNet}
    \begin{tabular}{cccccc}  
    \toprule
    Metric & w/o $\mathcal{L}_{pull}^2$ & w/o $\mathcal{L}_{push}^2$ & w/o CSD & w/o Denoising & Ours \\
    \midrule
    \textbf{mIoU} & 27.61 & 27.34 & 26.70 & 28.24 & \textbf{28.99} \\
    \textbf{mAcc} & 38.82 & 38.41 & 37.78 & 39.75 & \textbf{42.11} \\
    \bottomrule
    \end{tabular}
\end{table}

\subsection{Fine-grained capability of CLIP}

\noindent
To ensure a fair comparison with baselines, we used CLIP as the  semantic query tool for all samples. To assess CLIP’s reliability for segmentation, we compute the proportion of valid samples for each scene on the Lerf\_ovs dataset, as summarized in Table~\ref{tab:lerf_ovs_clip_success}. We find CLIP produces valid queries over 90\% of cases. This record is relatively lower in the kitchen scenes, where thin, reflective objects ({\it e.g.} spatulas and vessels) exhibit weak texture cues and are therefore harder to recognize. Future work can use better query tools ({\it e.g.} BLIP) to further improve the performance in such scenes. 

\begin{table}[h]
\centering
\renewcommand{\arraystretch}{1.2} 
\caption{CLIP query success rate on Lerf\_ovs dateset. We report the number and percentage of valid samples per scene under different scales.}
\resizebox{\linewidth}{!}{%
\begin{tabular}{c c c c c c}
\toprule
Scale & figurines & ramen & teatime & waldo\_kitchen & Overall \\
\midrule
whole       & 56/56 (100.0\%) & 62/71 (87.3\%) & 57/59 (96.6\%) & 14/22 (63.6\%) & 189/208 (90.9\%) \\
part        & 53/56 (94.6\%)  & 62/71 (87.3\%) & 57/59 (96.6\%) & 14/22 (63.6\%) & 186/208 (89.4\%) \\
whole+part  & 56/56 (100.0\%) & 64/71 (90.1\%) & 59/59 (100.0\%) & 16/22 (72.7\%) & 195/208 (93.8\%) \\
\bottomrule
\end{tabular}%
}
\label{tab:lerf_ovs_clip_success}
\end{table}

\subsection{Robustness to Error Accumulation via Feature Perturbation}
\label{sec:error_accumulation}

In cascaded optimization, errors from the first-stage clustering (e.g., imperfect global instance features) may propagate to later stages and affect local contrastive learning.
To quantify such error accumulation in a controlled manner, we conduct a noise-interference study by perturbing the global learned instance features before the first-stage clustering.
Let $\mathbf{f}_i\in\mathbb{R}^{C}$ denote the global instance feature of Gaussian $i$ right before first-stage clustering.
We construct a perturbed feature:
\begin{equation}
\tilde{\mathbf{f}}_i \;=\; \mathbf{f}_i \;+\; \tau\,\mathbf{z}_i,
\label{eq:noise_injection}
\end{equation}
where $\tau$ is a scalar controlling the perturbation strength and $\mathbf{z}_i$ is sampled from a zero-mean Gaussian distribution with per-channel scaling:
\begin{equation}
\mathbf{z}_i \sim \mathcal{N}\!\left(\mathbf{0},\,\mathbf{\Sigma}\right), \quad 
\mathbf{\Sigma}=\mathrm{diag}\!\left(\boldsymbol{\sigma}^2\right).
\label{eq:noise_distribution}
\end{equation}
Here $\boldsymbol{\sigma}\in\mathbb{R}^{C}$ is computed from the global features over all Gaussians in the scene as the per-channel standard deviation, so that $\tau$ represents a relative noise magnitude that is comparable across channels. We sweep $\tau\in\{0,0.05,0.10,0.15,0.20\}$, perform first-stage clustering using the perturbed features $\tilde{\mathbf{f}}_i$, and keep all other training/evaluation settings unchanged.
We then evaluate how upstream perturbations affect the subsequent local contrastive learning stage, reporting mIoU on the teatime scene.

\begin{figure}[h]
\centering
\includegraphics[width=\linewidth]{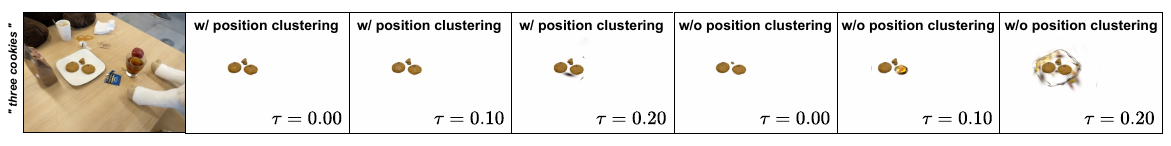}
\caption{\textbf{Qualitative robustness under feature perturbation.}
Visualization on \textit{teatime} at increasing noise magnitudes ($\tau\in\{0.00,0.10,0.20\}$).
Position-aware clustering produces more stable masks under stronger perturbations.}
\label{fig:cookie}
\end{figure}

\noindent\textbf{Position-aware clustering mitigates error propagation.}
As shown in Table~\ref{tab:noise_tau} and Fig.~\ref{fig:cookie},
when clustering is performed in a joint (feature + 3D position) space, the performance degrades more gracefully as $\tau$ increases.
For example, as $\tau$ grows from $0$ to $0.20$, mIoU drops from $53.68$ to $50.51$ with position-aware clustering, versus $48.60$ to $40.33$ without it.
This suggests that incorporating spatial cues stabilizes cluster assignments under feature corruption, thereby reducing error accumulation across cascaded stages.

\begin{table}[h]
\centering
\setlength{\tabcolsep}{12pt}
\caption{\textbf{Robustness to feature perturbation.}
We inject Gaussian noise of magnitude $\tau$ into global instance features before first-stage clustering and evaluate the downstream mIoU on teatime scene.
Position-aware clustering consistently yields higher mIoU and degrades more gracefully as $\tau$ increases.}
\label{tab:noise_tau}
\begin{tabular}{cccccc}
\toprule
$\tau$ & 0.00 & 0.05 & 0.10 & 0.15 & 0.20 \\
\midrule
w/ position clustering  & 53.68 & 53.10 & 52.57 & 51.77 & 50.51 \\
w/o position clustering & 48.60 & 47.89 & 45.88 & 42.71 & 40.33 \\
\bottomrule
\end{tabular}
\end{table}

\subsection{More Visualization}

\begin{figure*}[h]
\centering
\includegraphics[width=0.75\linewidth]{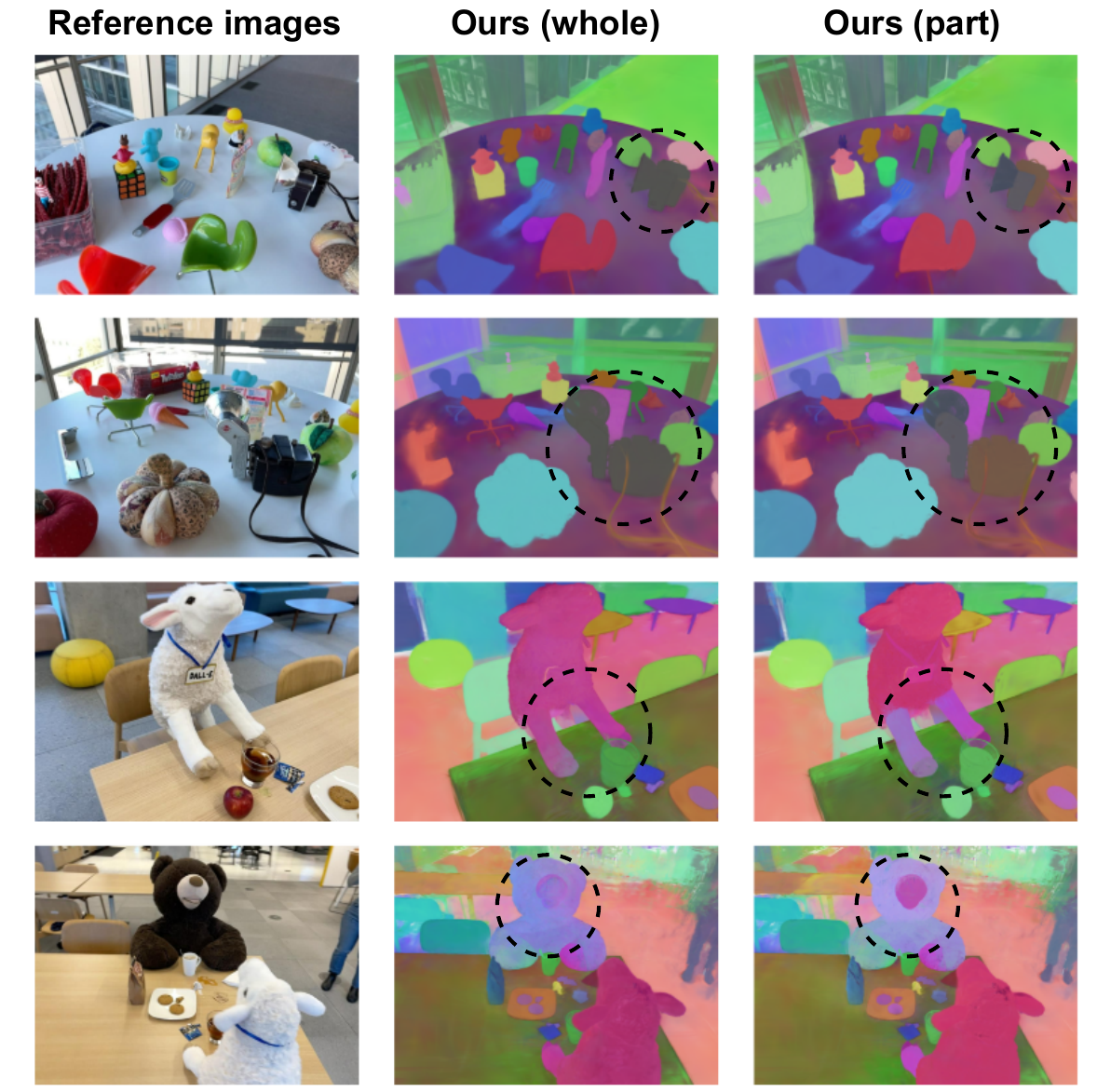}
\caption{More results with the rendered instance feature maps.}
\label{fig:Appendices2}
\end{figure*}

Figure~\ref{fig:Appendices2} presents the rendered instance feature maps for more multi-view images. Black‑circled regions highlight the consistency of the instance at different scales.

\begin{figure*}[h]
\centering
\includegraphics[width=1.0\linewidth]{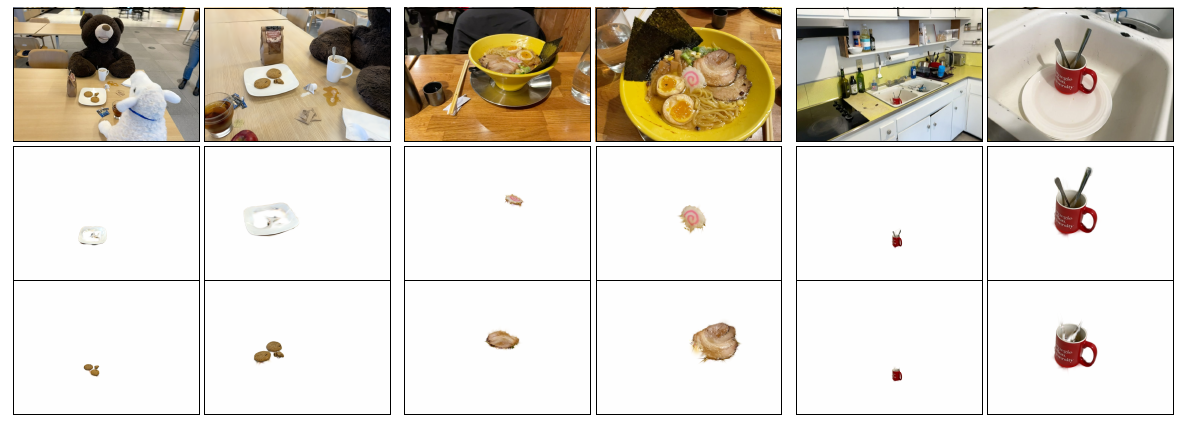}
\caption{More results with the rendered objects across different scenes.}
\label{fig:Appendices3}
\end{figure*}

\noindent
Figure~\ref{fig:Appendices3} illustrates more rendered objects across different scenes, demonstrating the effectiveness of our method.

\begin{figure*}[h]
\centering
\includegraphics[width=1.0\linewidth]{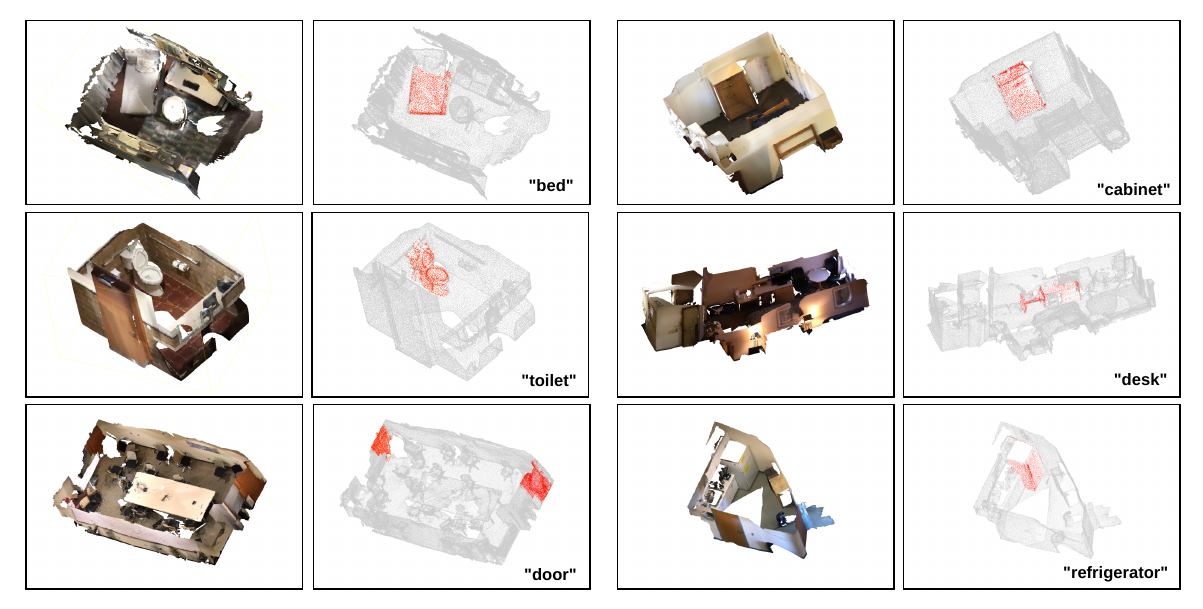}
\caption{More results of point cloud clusters, demonstrating the 3D point-level segmentation ability of our method.}
\label{fig:Appendices4}
\end{figure*}

\noindent
Figure~\ref{fig:Appendices4} illustrates more visualizations of 3D point cloud clusters on the ScanNet dataset, demonstrating the 3D segmentation ability of our method.

\begin{figure*}[h]
\centering
\includegraphics[width=1.0\linewidth]{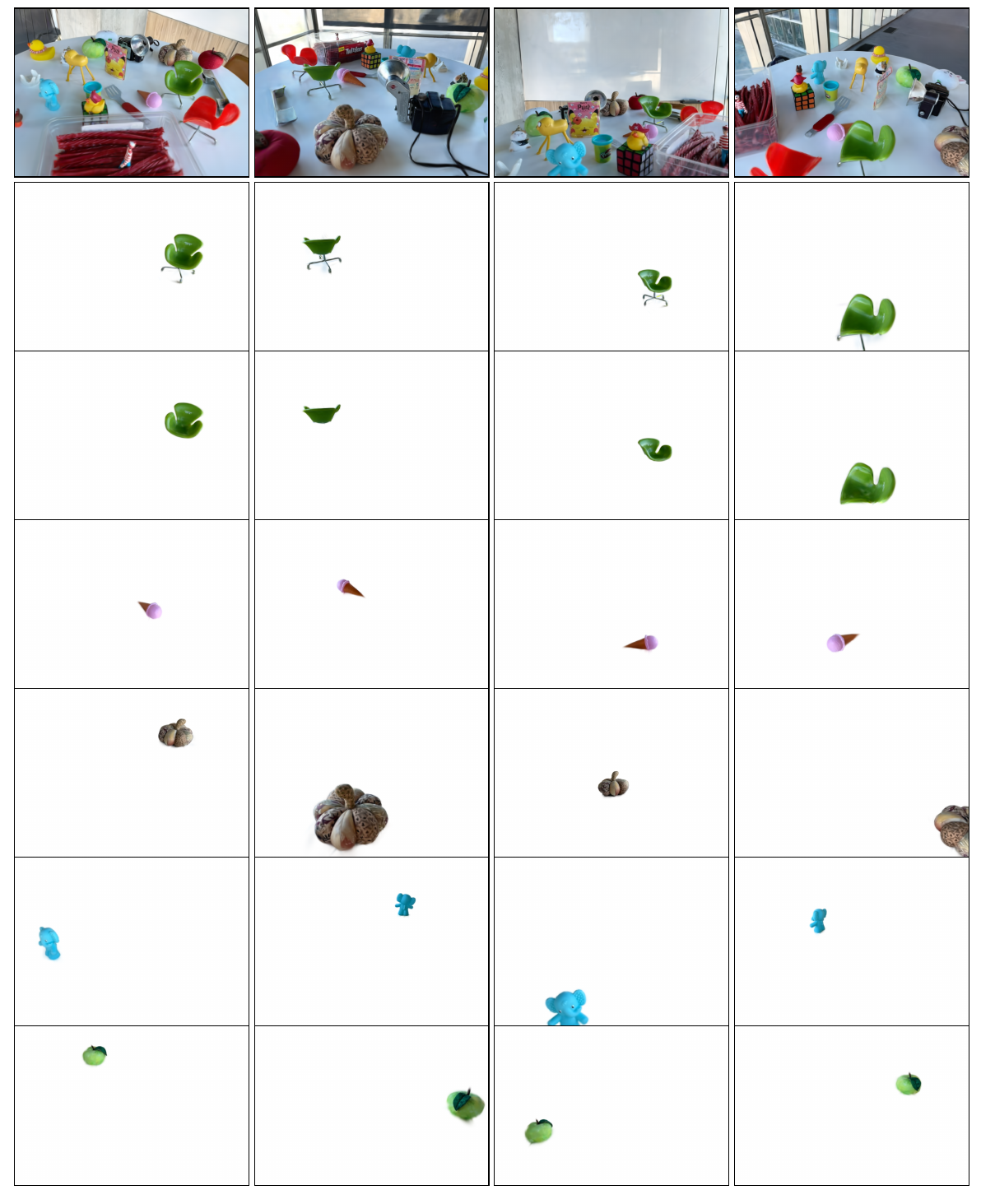}
\caption{More results with the rendered objects across different views, demonstrating the multi-view consistency of our method.}
\label{fig:Appendices5}
\end{figure*}

\noindent
Figure~\ref{fig:Appendices5} illustrates more rendered objects across different views, demonstrating the multi-view consistency of our method.

\end{document}